\documentclass[runningheads]{llncs}

\usepackage[T1]{fontenc}

\usepackage{graphicx}
\usepackage{amsmath}
\usepackage{amssymb}
\usepackage{xcolor}

\begin{document}
\title{Sequence Recognition in Bharatnatyam dance}
\author {Himadri Bhuyan \orcidID{0000-0002-5105-9390} \and 
Rohit Dhaipule \orcidID{0000-0003-2874-5517} \and
Partha Pratim Das  \orcidID{0000-0003-1435-6051} \\
\email{himadribhuyan@gmail.com},
\email{rohit1045d@gmail.com},
\email{ppd@cse.iitkgp.ac.in}}
%\and
%Third Author\inst{3}\orcidID{2222--3333-4444-5555}}
%
%
\authorrunning{H. Bhuyan et al.}
% \institute{IRIT Laboratory, Paul Sabatier University, Toulouse, France
% \email{\{max.halford,frank.morvan\}@irit.fr}\\ \and
% IMT Laboratory, Paul Sabatier University, Toulouse, France\\

\newcommand\acceptednote[1]{%
  \begingroup
  \renewcommand\thefootnote{}\footnote{#1}%
  \addtocounter{footnote}{-1}%
  \endgroup
}

\institute{Indian Institute of Technology Kharagpur, Kharagpur 721302, West Bengal, India  \\
\email{himadribhuyan@gmail.com}}

\maketitle              % typeset the header of the contribution
\acceptednote{The paper is accepted at the 7th International Conference on Computer Vision \& Image Processing (CVIP 2022). \url{https://doi.org/10.1007/978-3-031-31407-0\_30}}
\begin{abstract}
{\em Bharatanatyam} is the oldest Indian Classical Dance (ICD)
which is learned and practiced across India and the world. {\em Adavu} is the
core of this dance form. There exist 15 {\em Adavu}s and 58 variations. Each
{\em Adavu} variation comprises a well-defined set of motions and postures
(called dance steps) that occur in a particular order. So, while learning
{\em Adavu}s, students not only learn the dance steps but also take care of
its sequence of occurrences. This paper proposed a method to recognize
these sequences. In this work, firstly, we recognize the involved Key Postures (KPs) and motions in the {\em Adavu} using Convolutional Neural Network (CNN) and Support Vector Machine (SVM), respectively. In this, CNN achieves 99\% and SVM’s recognition accuracy becomes 84\%. Next, we compare these KP and motion sequences with the ground truth to find the best match using the Edit Distance algorithm with an accuracy of 98\%. The paper contributes hugely to the state-of-the-art in the form of digital heritage, dance tutoring system, and many more. The paper addresses three novelties; (a) Recognizing the sequences based on the KPs and motions rather than only KPs as reported in the earlier works. (b) The performance of the proposed work is measured by analyzing the prediction time per sequence. We also compare our proposed approach with the previous works that deal with the same problem statement. (c) It tests the scalability of the proposed approach by including all the {\em Adavu} variations, unlike the earlier literature, which uses only one/two variations.  

\keywords{Adavu \and Bharatanatyam \and Key posture \and Motion \and SVM \and CNN \and HOG \and MHI}
\end{abstract}
\section{Introduction}
%A dance performance consists of various information such as-- audio, video, temporal and spatial. Digitizing dance requires a systematic analysis and understanding of semantics. It can be useful in the several applications; dance interpretation, building tutoring system,  preservation of cultural heritage, and dance synthesis. However, it is not an easy task. There have been some works reported with Non-Indian Classical Dance like Ballet, Samba, Salsa. But very little work has been reported on Indian classical dance (ICD). 

{\em Bharathanatyam, Kuchipudi, Kathak, Manipuri, and Kathakali} are the most famous ICDs. {\em Bharatanatyam} is a significant ICD form with a perplexing mix of visual and audio data. To analyze {\em Bharatanatyam} dance, one must analyze {\em Adavu}, the basic building blocks. The {\em Adavu}s is the collection of postures and movements. There are 15 classes of {\em Adavu}s having 58 variations. 

\begin{figure}[!ht]
    \centering
    \includegraphics[scale=0.35]{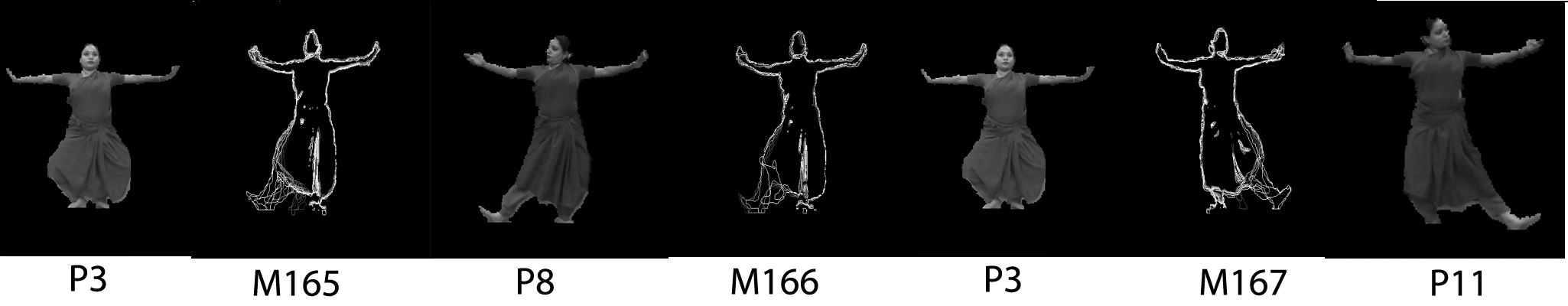}
    \caption{An example showing the sequence of KPs and motion in {\em Natta-1}}
    \label{fig:SequenceOFfeatures}
\end{figure}

The visual data comprises of key postures, motions, and trajectories. A key posture (KP) in which the dancer remains stationary for a while. A motion is a part of the dance from one KP to the accompanying KP. Motions (M) and KP occur alternately and may repeat in a performance. Figure~\ref{fig:SequenceOFfeatures} shows this. The audio data is known as {\em Sollukattu} which comprises of Bols, Beats, and tempo. In this work, we only deal with the visual information of the {\em Adavu}s.

%  \begin{figure}[!htp]
%     \centering
%     \includegraphics[scale=0.8]{figures/adavu-ex.png}
%     \caption{An example of Adavu}
%     \label{fig:adavuex}
% \end{figure}

In this work, we use the Key Posture and Motion information and train a model to recognize the {\em Adavu}s using Edit Distance algorithm. We first solve the sub-problems of recognizing the KP and Motion classes which will be used to recognize the Adavu class. 

The data set is created by \cite{himadri2021}, where the RGB, Depth, and Skeleton videos of dance performances have been captured using Kinect V1. The videos are then manually segmented into Key Frames (KF) sequences corresponding to the expected KPs and Motions. To recognize the KPs in the segmented KFs, we follow two approaches - SVM and CNN. SVM uses HOG as a feature between these two approaches, and CNN takes the raw KPs for classification. Similarly, for motions, we first merge all the key frames related to a particular motion and generate Motion History Images (MHI), from which HOG features are extracted. SVM uses these features to build the trained model. 

\section{Related Work}
The recognition of {\em Adavu} is based on recognizing the KPs and motions involved. So, the given problem is divided into three sub-problems that are be solved: (a)  KP recognition, (b) Motion recognition, and (c) {\em Adavu} recognition as a sequence of KPs and Motions. Hence a survey is conducted on the recent works related to the mentioned sub-problems.

\subsection{Recognition of Key Postures} \label{subsec:kprecogrelwork}
Over time, many authors have worked in the area of posture recognition. ~\cite{jain2021enhanced,biswas2021classification} classify eight different ICDs based on different postures as input and recognize the type of ICD to that posture belongs.  ~\cite{jain2021enhanced} use ResNet50 whereas, ~\cite{biswas2021classification} use and  VGG16 \& VGG19 (pre-trained) as classifier. A similar work is reported in ~\cite{naik2020classification}, which tries to classify five ICDs based on postures using CNN. In all these works, authors work on recognizing the body postures. Authors in ~\cite{shailesh2020computational} design a framework to classify different foot postures using Naive Bayes classifier.~\cite{mohanty2016nrityabodha} recognize the ICD postures using CNN. The data set comprises 26 postures where Kinect captures 12 postures in a  controlled environment, and 14 are taken from YouTube videos. ~\cite{saha2013gesture}, ~\cite{venkatesh2016automatic} include recognition of Key Postures in ICD using SVM and obtain good results.

CNN, SVM, and GMM were used by \cite{mallick2022posture} with RGB and Skeleton-Angle as input to recognize the KPs in {\em Bharatanatyam's} {\em Natta Adavu}s. The best result (99.56\%) has been obtained using ResNet (32 CV Layers, 2 FC Layers)  for 23 classes with RGB data, whereas SVM with RGB-HOG also does relatively well (97.95\%). They also used GMM with skeleton angle as features, resulting in an accuracy of  83.04\%.

Further, authors in the paper \cite{himadri2021} use SVM for KP recognition in ICD. The authors use two approaches to classify key postures. In the first approach, angles of skeleton bones are used as features, whereas in the second approach, HOG features from the RGB frames are used as features. In both the approaches, SVM has been used to recognize the Key postures and achieve 93.36\% and 94.15\% respectively. Using this model and the Edit Distance Algorithm, the {\em Adavu}s are recognized and classified. This has been performed only on two {\em Adavu}s,  {\em Natta} and {\em Mettu} and report 99.38\% accuracy. However, the authors do not consider motion aspects during this recognition.

%With the advancement in Deep Learning, many works like \cite{raman2022markerless}, \cite{burns2020using} have been published where CNN-based models were used to recognise the human poses. A PoseNet model, neural networks-based pose estimation method was proposed in \cite{yamao2021development}, and implemented by a Raspberry Pi to recognise the human poses. 17 key points on a human body are detected by the system and matched to the set of points registered beforehand. In \cite{raman2022markerless}, animal (particularly dog) pose recognition has been done using ResNet deep learning model. The results are shown on a dataset containing more than 5000 images of dogs in various poses. The system has been incorporated in a mobile app to track animals.

%In this paper, we use the data set created by Mallick \cite{mallick2022posture} and use SVM and CNN on RGB-HOG on KPs of all the 13 {\em Adavu}s of Bharatanatyam dance. 

\subsection{Recognition of Motion} \label{subsec:motionrecogrelwork}

Motion Recognition has been one of the most famous and worked problems in recent years. Worldwide many researchers propose different machine learning and deep learning techniques to solve it.

In \cite{chaudhry2017automatic}, with Spacetime interest point (STIP) and Histogram of optical flow (HOOF) as features, motion information has been extracted from the dance movements. KNN, Neural network, and Tree bagger classifier are explored. Here, the Treebagger classifier gives a maximum accuracy of 92.6\% with N=150 trees as a parameter. Similarly, in \cite{bhuyan2019motion}, the velocities of skeleton joints have been used as features, and methods like dynamic time warping (DTW) and the kNN algorithm are explored, resulting in the best accuracy of 85\%.

The growth of Deep Learning models has helped researchers work with the motions effectively. In \cite{kumar2018indian}, a new segmentation model is developed using discrete wavelet transform, and local binary pattern (LBP) features. This is combined with an AdaBoost multi-class classifier that was used to classify the dance actions. The Indian classical dance datasets used by the authors consist of performances on ‘Bharatanatyam’ and ‘Kuchipudi’ from online YouTube videos and offline dance videos recorded in a controlled environment at K.L. University, cams department studio.

Symmetric Spatial Transformer Networks (SSTN) were constructed by the authors of \cite{kaushik2018nrityantar} to recognize the ICD dance forms' sequences. They used the postures as features based on skeletal joints and 3D CNN as a classifier. In ~\cite{kishore2018indian} they use CNN to recognize or classify dance actions based on posture.  

Motion History Image (MHI) has been a simple, effective, and robust approach to recognizing human actions. Several works based on MHI have been reported in the article \cite{ahad2012motion}. In this paper, we use HoG of MHI as a feature in motion recognition.

%Instead of using the MHI in it's original form, authors of papers \cite{huang2011human}, \cite{tsai2015optical}, \cite{meng2009motion} have used different histogram techniques such as Histogram of Gradient (HoG), Histogram of Optical Flow (HoF), Motion history histogram (MHH) and then used ML models which resulted in a good accuracy.

%In this paper, we use the data set as in \cite{mallick2022posture} and use SVM on MHI-HOG of Motions of Bharatanatyam dance.

\subsection{Recognition of {\em Adavu}s} \label{subsec:adavurecogrelwork}

While performing an {\em Adavu}, the dancers follow a predetermined sequence and mix of Postures and Motions in sync with the musical beats. The dancer needs to ensure that the rules adapted to change from a KP to a Motion and back to another KP are followed. The classifier we build needs to ensure those rules are adhered to while recognizing a particular {\em Adavu}.

One of the first works on {\em Adavu} recognition ~\cite{sharma2013recognising} was reported based on the posture sequence, without considering the temporal information. Authors of ~\cite{kale2015bharatna} used RGB-D data which was captured using Kinect to recognize the Adavu. They get 73\% recognition accuracy. Later, authors of ~\cite{mallick2022posture} used RGB-HOG data in Hidden Markov Model (HMM) to recognize only 8 {\em Adavu}s, resulting in an accuracy of 94.64\% which is a significant improvement when compared to \cite{kale2015bharatna}. Recently, in ~\cite{himadri2021}, the authors use the posture information to recognize all the variations of {\em Mettu} and {\em Natta Adavu}s. In this, they recognize the occurrence of the postures using SVM in a given {\em Adavu} and use the Edit Distance algorithm to find the best match among the {\em Adavu} dictionary. Here, 12 {\em Adavu}s are considered and achieved an accuracy of 99\%.

The {\bf research gaps} identified in the literature survey are summarised as: (a) The earlier works~\cite{sharma2013recognising} ~\cite{mallick2022posture} ~\cite{himadri2021} ~\cite{kale2015bharatna} consider only KPs in the {\em Adavu} recognition. However, the involved motions are not included. It may be due to the complexity of the motion recognition. (b) Most of the earlier works~\cite{sharma2013recognising} ~\cite{mallick2022posture} ~\cite{himadri2021} ~\cite{kale2015bharatna} include only one or two {\it Adavu}s. However, there exist 58 {\em Adavu} variations. (c) Instead of the classical ML technique, an effective deep learning approach is yet to be explored. (d) The time complexity of the different approaches is yet to be analyzed.

In this paper, we address the above gaps as the {\bf contribution of the paper}:  (a) The {\em Adavu} recognition considers both motions and postures rather than only postures. (b) We take 13 {\it Adavu}s and 51 sub-classes into consideration which tests the robustness of the approach. (c) The paper designs an effective CNN which takes less time (d) The performance of the proposed approaches is analyzed based on the time complexity.

\section{Data Set} \label{Sec:DataSet}
We record all the {\em Adavu}s using Microsoft Kinect V1~\cite{kinect360} in a controlled environment as followed in~\cite{aich2017nrityaguru}. The Kinect captures three streams of data; RGB, Depth, and Skeleton. We use the RGB stream. 
The paper tries to recognize the {\it Adavu}s by considering both the KP and motion aspects. So, the dataset should include both of these attributes. Table~\ref{tab:AdavuRecogDataset} shows the {\it Adavu}, its variants, the involved KPs, and motions. Though there exist 15 {\em Adavu}s and 58 variations, by discarding the erroneous data (error in data extraction), the entire data set boils down to 13 {\em Adavu}s and 51 variations. There are 184 KPs and 334 motion classes. However, all the motions reported in the Table ~\ref{tab:AdavuRecogDataset} are not used due to their sample size. We only consider the motions having \# of samples more than 12 to avoid the biases of the motion recognizer. Similarly, Two KPs (P67 \& P69) are ignored. So, 182 classes of KPs are considered for recognition. ~\cite{AdavuFile} shows the KPs and its Train-Test split. Table~\ref{tab:AdavuRecogDataset} shows one of the annotations as an example. It gives information about the duration of the occurrence of the motions and KPs in a given {\it Adavu}.  % and 25. The behaviour of the classifiers with these data is also explored.
\begin{table}[!ht]
    \centering
    \scriptsize
    \caption{Involved KPs \& Motions in the {\em Adavu}s and Annotations }
    \begin{tabular}{|c|c|c|c|c|c|} \hline
       {\em Adavu}  & Variant & Total KPs & Unique KPs & Total Motions & Unique motions \\ \hline \hline
        Joining &3 &1350 &13		&78	&14 \\ \hline
        Katti\_kartari &1 &258 &2	&23	&4 \\ \hline
        Kuditta\_Nattal&6 &5861 &28	&231	&51 \\ \hline
        Kuditta\_tattal&5 &3914 &32	&616	&61 \\ \hline
        Mandi &2 &4525 &18		&192	&27 \\ \hline
        Mettu	&4 &4573 &25	&238	&32 \\ \hline
        Natta	&8 &9449 &19	&410	&54 \\ \hline
        Paikkal	&3 &1620 &12	&72	&14 \\ \hline
        Sarika	&4 &2720 &8	&191	&25 \\ \hline
        Tatta	&8 &6920 &1	&264	&2 \\ \hline
        TeiTei\_Dhatta &9 &1262 &9	&96	&15 \\ \hline
        Trimana	&3 &1973 &14	&149	&31 \\ \hline
        Uttasanga &1 &278 &3	&24	&4 \\ \hline \hline
        Total	&51 &44703 &184	&2584	&334 \\ \hline
         \multicolumn{6}{c}{} \\
         \multicolumn{3}{c}{\bf (a) Motion annotation}& \multicolumn{3}{c}{\bf (b) KP annotation}\\ \hline
        Motion ID  &Start-Frame  &End-Frame & KP ID  &Start-Frame  &End-Frame \\ \hline \hline
       M219 &103 &123 & P134  &124  &134  \\ \hline
       M220 &135 &205 & P135  &206  &225 \\ \hline
       M221 &226 &307 & P136  &308  &330 \\ \hline
       M222 &331 &403 & P28  &404  &416 \\ \hline
      .. &.. &.. &.. &.. &.. \\ \hline
      M224 &701 &780 &  P32  &781  &805 \\ \hline
        
    \end{tabular}
    
    \label{tab:AdavuRecogDataset}
\end{table}

\subsection{Data Set for {\em Adavu}}
A particular {\it Adavu} includes a sequence of motions and KPs as addressed in the annotation, Table~\ref{tab:AdavuRecogDataset}. Each KP and motion comprises a set of frames. The set of frames involved with KPs is momentarily stationary. However, in motion, these frames are dynamics. To identify an {\it Adavu}, we need to recognize the KPs and motions that occur in a definite sequence. For a KP, there exist several candidate key frames (KP implies a set of momentarily stationary frames). However, for motion, it is not. So, for a given {\it Adavu}, we can have several sequences of KPs and motions.  

%Note that each KP/Motion corresponds to a certain number of frames in the video. Therefore, there  are actually many candidate key frame subsequences  that can define any Adavu (where a KF is chosen from the set of KFs for a specific KP).

Let us consider {\it Paikkal-1 Adavu} as an example to construct such candidate sequences. Note that a set of motion frames gives rise to a single motion. However, each key frame is treated as KP in a video. So we can not count the number of valid sub-sequences based on motions. The number of valid sub-sequences will be restricted by the KP containing the smallest number of KFs. For {\it Paikkal-1 Adavu}, there will be 11 valid sub-sequences, which are restricted by 'P134' (see Table \ref{tab:AdavuRecogDataset}) which has only 11 frames. That means $n= min($\# of frames of a $KP \in$ given Adavu $)= 10$. Therefore, the $1^{st}$ sequence is: M219 -- P124 -- M220 -- P206 -- M221 -- P308 -- M222 -- P404 -- .... -- M224 -- P781 and the last sub-sequence of {\it Paikkal-1} is: M219 -- P134 -- M220 -- P216 -- M221 -- P318 -- M222 -- P414 -- ......... -- M224 -- P791. Using this strategy, we create sub-sequences from the videos in the data set that can be used as data instances for the classifiers. The number of possible sequences associated with each {\em Adavu} variant is shown in Table~\ref{tab:NumberofSequnces}. Here, we should remember that a given sequence always starts with a KP. Before starting the dance, the dancer takes a random posture which denotes that the dancer is ready to dance, but this start KP may not appear further in the given sequence. So, during the sequence annotation, we discard this start KP.
 \begin{table}[!ht]
     \centering
     %\tiny
     \caption{The number of Sequences associated with each Adavu variant}
   \resizebox{\textwidth}{!}{  \begin{tabular}{cc||cc||cc||cc||cc||cc} \hline
         Adavu	&Sequneces		&Adavu	&Sequneces &Adavu	&Sequneces &Adavu	&Sequneces &Adavu	&Sequneces &Adavu	&Sequneces \\ \hline \hline
Joining\_1	&33	&Tatta\_6	&67	&Natta\_7	&38	&Kuditta\_Nattal\_4	&11	&Mandi\_2	&11	&Tei\_Tei\_Dhatta\_2	&50	\\ \hline
Joining\_2	&55	&Tatta\_7	&70	&Natta\_8	&24	&Kuditta\_Nattal\_5	&13	&Paikkal\_1	&48	&Tei\_Tei\_Dhatta\_3	&16	\\ \hline
Joining\_3	&23	&Tatta\_8	&22	&Mettu\_1	&33	&Kuditta\_Nattal\_6	&58	&Paikkal\_2	&44	&Tirmana\_1	&17	\\ \hline
Kartari\_1	&63	&Natta\_1	&56	&Mettu\_2	&34	&Kuditta\_Tattal\_1	&13	&Paikkal\_3	&44	&Tirmana\_2	&13	\\ \hline
Tatta\_1	&79	&Natta\_2	&66	&Mettu\_3	&25	&Kuditta\_Tattal\_2	&20	&Sarika\_1	&15	&Tirmana\_3	&9	\\ \hline
Tatta\_2	&90	&Natta\_3	&45	&Mettu\_4	&19	&Kuditta\_Tattal\_3	&13	&Sarika\_2	&34	&Utsanga\_1	&25	\\ \hline
Tatta\_3	&60	&Natta\_4	&41	&Kuditta\_Nattal\_1	&10	&Kuditta\_Tattal\_4	&5	&Sarika\_3	&16	&	&	\\ \hline
Tatta\_4	&41	&Natta\_5	&36	&Kuditta\_Nattal\_2	&11	&Kuditta\_Tattal\_5	&21	&Sarika\_4	&12	&	&	\\ \hline
Tatta\_5	&20	&Natta\_6	&30	&Kuditta\_Nattal\_3	&10	&Mandi\_1	&17	&Tei\_Tei\_Dhatta\_1	&19	&Total	&1645	\\ \hline
     \end{tabular}
     }
     \label{tab:NumberofSequnces}
 \end{table}
 \section{Workflow}\label{Sec:AdavuRecogworkFlow}
\begin{figure}[!ht]
    \centering
    \includegraphics[scale=0.3]{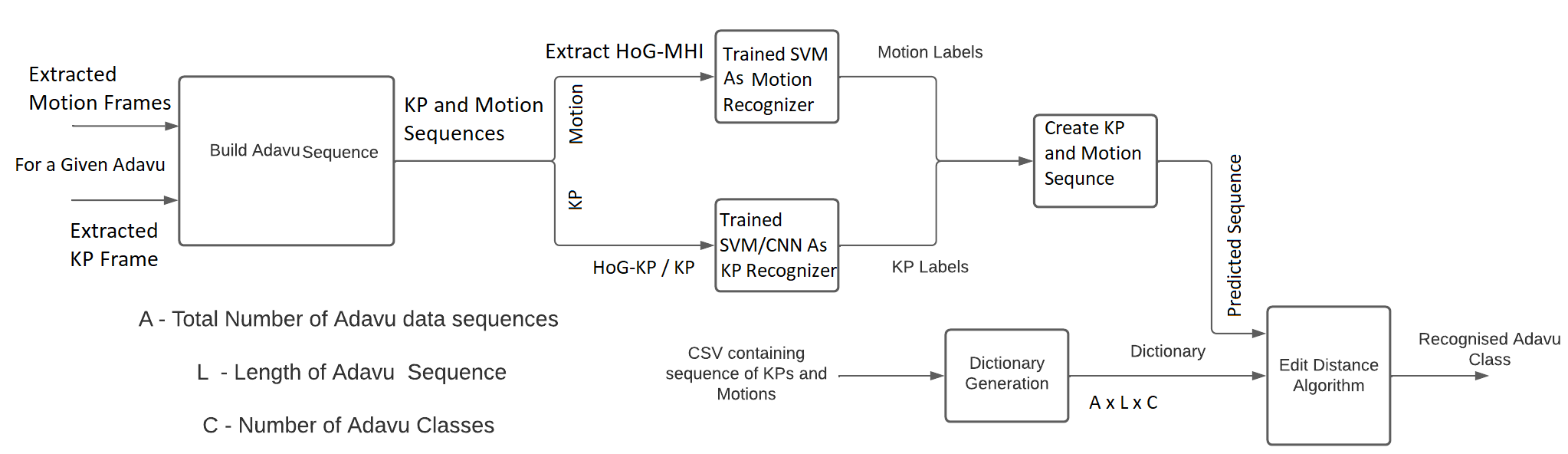}
    \caption{{\em Adavu} Recognition Workflow}
    \label{fig:AdavuRecogworkflow}
\end{figure}

Figure~\ref{fig:AdavuRecogworkflow} shows the workflow of the proposed approach. In this recognition process, We give an {\it Adavu} video as an input along with the annotation file (Refer Table~\ref{tab:AdavuRecogDataset}). We convert each video frame into a gray frame using the equation $GrayFr_i(x,y) = 0.299*Fr_i(x,y)_{Red} +0.587*Fr_i(x,y)_{Green} + 0.144*Fr_i(x,y)_{Blue}$ and subtract the background using the binary mask, $D_i(x,y) = Fr_i(x,y) * Mask_i(x,y)$  that can identify the dancer in a frame using depth information (captured by Kinect v1) as reported in~\cite{bhuyan2021automated}.  Where, $Mask_i(x,y) = 1$ if there exist dancer in  $Fr_i(x,y)$ otherwise $Mask_i(x,y) = 0$.

% Eq.~\ref{Eq:RGB_gray} and subtract the backgorund using the binary mask (Refer Eq.~\ref{Eq:BackgroundRemove}) that can identify the dancer in a frame using depth information as reported in~\cite{bhuyan2021automated}, which is captured by Kinect v1.

% \begin{equation}\label{Eq:RGB_gray}
% \begin{split}
%     GrayFr_i(x,y) &= 0.299*Fr_i(x,y)_{Red} \\ &+0.587*Fr_i(x,y)_{Green} \\ &+ 0.144*Fr_i(x,y)_{Blue}
%     \end{split}
% \end{equation}
% Where $i$ denotes the frame number and $Fr_i(x,y)$ is the intensity value of the frame $Fr_i$ at the location $(x,y)$. 

% \begin{equation} \label{Eq:BackgroundRemove}
%     D_i(x,y) = Fr_i(x,y) * Mask_i(x,y)
% \end{equation}
% Where 
% $ Mask_i(x,y) = \begin{cases} 
% 1, & \text{if there exist dancer in } Fr_i(x,y) \\
% 0, & \text{Otherwise}
% \end{cases}
% $

%  \begin{figure}[!ht]
%     \centering
%         \includegraphics[scale=0.45]{figures/gen_workflow.png}
%         \caption{Showing the {\em Adavu} recognition Work Flow}
%     \label{fig:AdavuRecogworkflow}
% \end{figure}

After the grayscale conversion and background removal, we extract the feature for KPs and motions that occur in the given video separately. The feature extraction is discussed in Section~\ref{sec:ExtractFeature}. Using the required feature, our trained ML model identifies the KP and motion in the {\em Adavu}. For KP recognition, CNN uses background subtracted frames and automatic feature extraction in convolutional layers. Whereas, for motion, the HOG-MHI feature is extracted by us and given as an input to SVM for classification. The recognized KP and motion sequences are fed to the {\em Adavu} Recogniser to recognize the {\em Adavu}. We use Edit distance for this {\em Adavu} classification.

%For KP recognition, we use HOG as feature and SVM as classifier. Whereas, for motion, the feature and classifier are HOG-MHI and SVM respectively. The recognised KP and motion sequences are fed to the {\em Adavu} Recogniser to recognise the {\em Adavu}. We use, Edit distance for this {\em Adavu} classification. 

\section{Feature Extraction} \label{sec:ExtractFeature}
The resultant frames are used for feature extraction after the grayscale conversion and background removal. The frames are of size $480 \times 640$ given as an input to compute the feature. We can use the annotation to distinguish KP and motion frames in a given {\em Adavu}. As we use CNN for KP recognition, CNN takes care of feature extraction in its convolutional layers. So, here we discuss only the motion feature extraction technique.

% \subsection{Feature extraction for KP} \label{sec:generateHOG}

% The KP frame with the size $480 \times 640$ is resized to  $120 \times 160$. Now, we compute the gradient of each pixel ($Dfx_i(x,y)$ and $Dfy_i(x,y)$) in a frame, $Fr_i$ using the intensity values in each pixel $(x,y)$ as $Dfx_i(x,y) = Fr_i(x+1) -  Fr_i(x-1)$ and $Dfy_i(x,y) = Fr_i(y+1) -  Fr_i(y-1)$. The gradient's magnitude ($GM$) and direction ($GDir$) are computed as $GM = \|Dfx_i(x,y), Dfy_i(x,y)\|$ and $GDir = Dfy_i(x,y)/Dfx_i(x,y)$ respectively. 

% After obtaining the gradient of each pixel, the gradient matrices (magnitude and angle matrix) are divided into $8\times8$ cells to form a block. For each block, a 9-point histogram is calculated with $GDir$ is used as bin and $GM$ as its value. Summing the values in each bin, we get a vector of size 9 (as there are 9 bins). Now a window is created by 4 adjacent blocks (i.e. $2\times2$) whose vectors are concatenated together to form a resultant vector of size 36 (= $4\times9$).  Now this concatenated vector is computed by moving the window to the right in each row. All these vectors are then concatenated to form a HoG feature vector of size 9576 (= \( 36\times(\frac{120}{8}-1)\times(\frac{160}{8}-1) \)). The SVM uses this feature to classify the KPs. 

\subsection{Feature Extraction for Motion} \label{Sec:Motionfeature}
First, for the motion frames, a matrix MHI of size $480 \times 640$ is initialized to zeros. After that, (a) The contours of each gray-scale frame are computed, (b) the Difference of contours between the two frames is computed, (c) A threshold will be applied now to create a binary image, and (d) The MHI will be updated with the values of differential binary frames, and (e) This process continues until the last frame of a particular motion instance resulting in an MHI. Figure~\ref{fig:SequenceOFfeatures} shows the motion MHI of M165, M166, and M167.

% \begin{itemize}
%     \item The contours of each gray-scale frame are computed. 
%     \item Difference of contours between the two frames are computed.
%     \item A threshold will be applied now to create a binary image.
%     \item The MHI will be updated with the values of differential binary frames.
%     \item This process continues until the last frame of a particular motion instance resulting in an MHI. Figure~\ref{fig:SequenceOFfeatures} shows the motion MHI of M165, M166, and M167.
% \end{itemize}

After the above process, we resize the MHI image from $480 \times 640$ to $120 \times 160$ and compute the HOG~\cite{dalal2005histograms} resulting in a feature vector of size 9576. These features will be used for Motion Recognition.

%  \begin{figure}[!ht]
%     \centering
%     \includegraphics[width=7cm, height=5cm]{figures/mhi_m70.png}
%     \caption{MHI of motion M71}
%     \label{fig:mhiex}
% \end{figure}

%For Adavu Recognition, as we know that each {\em Adavu} has a sequence of KPs and Motions. Each KP and Motion in the sequence will be replaced by the corresponding HOG features computed earlier resulting in the sequence of HOG features. For example, consider the sequence of Paikkal-1 Adavu, M1 - P1 - M2 - P2 - M3 - P3 - M4 - P1 - M3 - P4 - M2 - P3. This sequence contains 6 KPs and 6 Motions. On replacing the HOG features in this sequence we get a matrix of size $12 \times 9576$ which will be used for Adavu Recognition.

\subsection{Feature for {\em Adavu}} \label{sec:AdavuFeature}

An {\em Adavu} comprises two basic attributes; KP and motion. The KPs and motions appear in a definite order within a given {\em Adavu}. We call it as {\em Adavu} sequence. To recognize an {\em Adavu} we need to identify the KPs and motions involved. That means, in an {\em Adavu}, we will have the sequence of the feature of KPs and motions. From these features, our machine learning (ML) models will predict the KPs and motions that occur in an {\em Adavu}. For example, Let us consider a segment of {\em Natta-1 Adavu}. It comprises of the KPs and motions as: P3 -- M165 -- P8 -- M166 -- P3 -- M167 -- P11 ( Refer Fig.~\ref{fig:SequenceOFfeatures}). A given sequence contains $|P|$ and $|M|$ number of KPs and Motions, respectively. On replacing the features ($f$) in this sequence we get a matrix of size $|P| \times |M| \times f$. Now, this feature is given as an input to the classifiers to recognize the individual KPs and motions, generating a sequence. This sequence helps us to recognize an {\em Adavu}.

\section{KP Recognition} \label{sec:KPRecog}
\begin{figure}[!ht]
    \centering
    \includegraphics[scale=0.4]{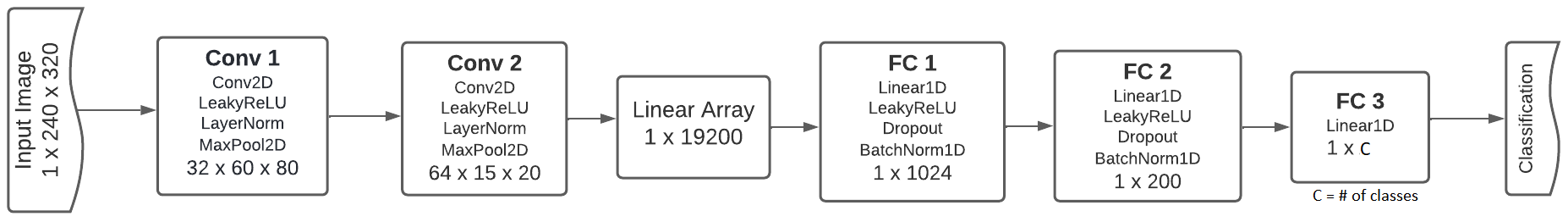}
    \caption{CNN architecture for KP Classification}
    \label{fig:KPCNNArchitect}
\end{figure}

This section discusses KP recognition, one of the attributes in the {\em Adavu}. Here, we use CNN as classifier and background subtracted frame as input. CNN compute the feature of each KP in its feature extraction layer. Input to the CNN is the background subtracted KP. Figure~\ref{fig:KPCNNArchitect} shows the the CNN architecture used for KP Recognition. The given CNN uses two convolutional (CV) layers and three fully connected (FC) layers.

We use 32 and 64 channels in two CV layers, respectively, with required padding in the input. We do apply layer normalization to normalize the feature. Finally, max-pooling is applied with the window $2 \times 2$ and stride two. The output of each CV layer is shown in the Figure~\ref{fig:KPCNNArchitect}

The operations involved in each FC layer are shown in Figure~\ref{fig:KPCNNArchitect}. To handle the overfitting, we apply {\em drop out} and limit the number of epochs to 15. The decision on the number of the epoch is taken by analyzing the train test accuracy (beyond 15, it is moving towards overfitting). Both techniques help prevent CNN from memorizing the data and driving towards learning. We use {\em Leaky ReLu} instead of {\em ReLu} to prevent the "Dead ReLu", since no learning happens for negative input in {\em ReLu}. Moreover, as reported in~\cite{dubey2019comparative}, Leaky ReLu also performs better than {\em ReLu}. The last FC layer's output is 182 feature arrays corresponding to 182 KP classes. These features are converted into class scores using the {\em softmax} layer and label the unknown KPs. 

Using CNN, out of 8943 test samples 8895 predicted correctly, that is, we get a recognition accuracy of 99.46\% along with the average Precision = 99.95\%, Recall = 99.15\% and F1 Score = 99.51\%. This good F1 Score shows the robustness of the result and classifier. Moreover, 158 KPs achieve 100\% accuracy and the rest score more than 95\% except for a few. Table~\ref{tab:KPresultCNN} shows the test result and misclassification details.

We also try the two popular CNN models, AlexNet~\cite{krizhevsky2012imagenet} and VGG16~\cite{russakovsky2015imagenet} for the KP recognition to learn the behavior of the models with different numbers of CV and FC layers. Finally, we reach our proposed CNN model with two CV and three FC layers. With the increase in depth (more number of CV and FC layers) the execution time for training and testing increases~\cite{he2015convolutional}. The execution time for training and testing is shown in Table~\ref{tab:DifferentCNN}. Again, after looking at the accuracy, precision, recall, and F1 score, we reported our CNN model as the best performer on the given data set. Table~\ref{tab:DifferentCNN} shows the result comparison among the proposed model, AlexNet,  VGG16, and ResNet. The robustness of the proposed CNN architecture is yet to be tested with the different types of data.

\begin{table}[!ht]
    \centering
    \caption{Result \& Time Comparison of different CNN Models}
    \label{tab:DifferentCNN}
    \begin{tabular}{c|c|c|c|c|c|c|c} \hline
    CNN Type	&Precision &Recall &F1 Score &Accuracy	&\# of KPs with & Training & Test \\
		&(\%)	&(\%)	&(\%)	&(\%)	& Accuracy = 100\% &Time(s) &Time(s)\\ \hline
AlexNet~\cite{krizhevsky2012imagenet}		&79.62	&80.35	&79.14	&94.14	&99 &7415.60 &41.97 \\ \hline
VGG16~\cite{russakovsky2015imagenet}		&90.06	&88.54	&88.86	&95.73	&92 &9393.10 &72.10 \\ \hline
ResNet~\cite{mallick2022posture}  &99.27  &98.77  &98.97  &99.54  &164  &6410.02 &58.43 \\ \hline
Our Model	&99.95	&99.15	&99.51	&99.46	&158 &4672.75 &36.37 \\ \hline
    \end{tabular}
    
\end{table}

\subsection{KP Recognition: Result Discussion}
Table~\ref{tab:KPresultCNN}  shows the result. CNN achieves 99.46\% accuracy. In this, only 48 samples are mispredicted out of 8,895 test samples.

\begin{figure}
    \centering
    \includegraphics[scale=0.35]{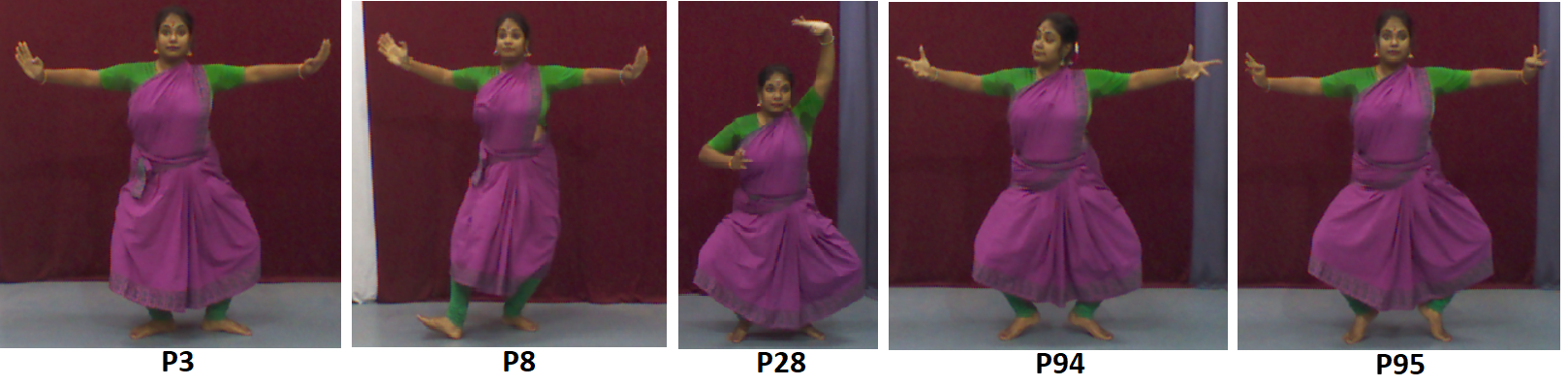}
    \caption{Most misclassified KPs-- P3, P8, P28, P94 and P95 as reported in CNN}
    \label{fig:KPs3_8_28_94_95}
\end{figure}

Most of the misclassifications are well expected. In P8, the dancer stretches their right leg. When these stretches are insignificant, it resembles P3 just as bending. So, in some cases, P8 is mispredicted as P3. In P94 and P95, postures are very much similar. The only difference is there in the leg position. In P94, the foot is grounded, whereas, in P95, the dancer stands on the toes. Though P28 seems very dissimilar from P3, some samples of P28 are mispredicted as P3. Sometimes in P3 and P28, the dancer's hand could not be tracked by Kinect while stretching the left hand towards the left or raising the hand upward. So, in that scenario, P28 becomes similar to P3 and gets mispredicted. Figure~\ref{fig:KPs3_8_28_94_95} shows the mispredictions KPs. 

\begin{table}[!ht]
    \centering
    \tiny
    \caption{KP Recognition Test Result using CNN}
    \begin{tabular}{cccc||cccc||cccc} \hline
         Class  &Test 	 	&Accuracy  &Miss	&Class  &Test 	 &Accuracy  &Miss &Class  &Test 	 &Accuracy  &Miss\\ 
\#&Sample  		&(\%)	   & Predictions        &\#	&Sample  &(\%)	    & Predictions &\#	&Sample  &(\%)	    & Predictions\\ \hline \hline
P3	&1134	&99.91	& P11(1)	&P46	&110	&99.09	&P3(1)	&P89	&5	&80	& P183(1)	\\ \hline
P8	&413	&98.06	& \textcolor{red}{P3(8)}	&P49	&46	&97.82	& P48(1)	&P94	&14	&57.14	&\textcolor{red}{P95(6)}	\\ \hline
P11	&406	&99.75	& P3(1)	&P52	&67	&98.50	&P3(1)	&P100	&161	&99.37	&P103(1)	\\ \hline
P23	&40	&97.5	& P52(1)	&P74	&12	&91.66	& P60(1)	&P103	&111	&99.09	& P100(1)	\\ \hline
P26	&16	&93.75	& P17(1)	&P75	&9	&88.88	& P54(1)	&P109	&33	&96.96	&P125(1)	\\ \hline
P28	&47	&87.23	&\textcolor{red}{P3(6)}	&P78	&58	&98.27	& P76(1)	&P163	&10	&90	& P3(1)	\\ \hline
P42	&14	&92.85	& P46(1)	&P79	&154	&99.35	&P77(1)	&P173	&64	&95.31	& P6(1), P3(2)	\\ \hline
P43	&43	&97.67	& P48(1)	&P84	&44	&95.45	& P3(2)	&P181	&12	&91.66	& P52(1)	\\ \hline

\multicolumn{12}{c}{ * The KP classes attending $<$ 100\% accuracy are reported in this table}\\
    \end{tabular}
    \label{tab:KPresultCNN}
\end{table}

\section{Motion Recognition} \label{sec:MotionRecog}
In {\em Adavu} each KP is followed by a motion. To recognize an {\em Adavu}, we also need to recognize the motion. Here, we use SVM to recognize the motion. SVM uses HoG of MHI as an input for the classification~\cite{MotionRecogDance2022}. The dance motion recognition comes up with several challenges, such as (a) Motions' complexity, (b) Occlusion due to the dressing style that is being followed in any Indian classical dance, and (c) Uneven frames involved with each motion. Even the same motion may not have the same number of frames, making it challenging to generate an equal-sized feature vector for any ML model. We try to handle these challenges and can classify the motion.

As discussed in section~\ref{Sec:Motionfeature}, we create MHI for a given motion which is a 2D vector of size $120 \times 160$. Now, this MHI is given as an input to the HOG generator to generate the HOG of MHI image. For each motion, it yields a feature vector of size 9576 (Refer Section~\ref{sec:ExtractFeature}). It is used as a feature in SVM.

We use SVM~\cite{chang2011libsvm} of One-Vs-One for this classification. Here, we choose only those motion classes with samples $\geq 12$, since most of the motions have very few samples. By this, 54 motion classes are left; that is, out of 334 motions, only 54 are chosen. Of 320 motion samples, 271 predicted correctly, resulting in an overall 84.08\% accuracy. In most of the classes, one or two samples are wrongly predicted. 

\subsection{Motion Recognition-- Result Discussion} \label{sec:ResultdiscusionMotion}
During motion recognition, following observations are made:
\begin{itemize}
\item The quantum of misclassifications is not high since most of the mispredictions occur for one or two samples. The mispredictions of zero, one, and two/three are 22, 20, and 12, respectively. That implies, 22 motion classes reach 100\% accuracy. 

    \item The motions having similar characteristics are getting misclassified, which is well expected. For example, M72 as M73, M74 as M75, and M75 as M74. In M72, with the help of toes, the body is raised, and then the body comes down. However, In M73, the body weight is given to the right foot, and the left foot is tapped. So, the MHI of M72 becomes very close to M73, and some samples of M72 are misclassified as M73. A similar pattern is observed between M74 and M75. Here, M74 and M75 are mirror images of M72 and M73, respectively.

\item Sometimes, motion in the leg remains unrecognizable due to occlusion (the dress in the lower body part). It implies that the MHI can not give us the expected result and may be a reason for mispredictions. For Example, M110 is misclassified as M112 and vice versa.

\item There is misclassification of the motions M287 \& 86 and M244 \& M334. In M287, the dancer takes a half circular anti-clockwise movement from a side view (the left side towards the camera), faces the camera, and then taps the left foot. In contrast, the M86 motion is only taping the left foot. Here, the half circular motion around the torso may not be recorded by MHI, making M287 and M86 the same kind of motion. 

M244 and M334 are minimal motion, that is, these motions occur for a very small number of frames. In M244, the dancer stands on the heels making tows up, and then taps the tows on the floor. However, M334 is just opposite M244. In this, dancer stands on its tows by taking the body slightly up and then comes down by striking the heel on the floor. These two motions are hardly distinguishable and get misclassified. The MHI of the feet is hardly recognized. However, the slight change of MHI is visible on the hand due to the vertical movement of the body. %Refer the MHI Figure~\ref{fig:M287_86nM244_334}.

%     \begin{figure}[!htb]
%     \centering
%     \includegraphics[scale=0.95]{figures/M287_86n M244_334.png}
%     \caption{Showing MHI of M287 \& 86 and M244 \& 334}
%     \label{fig:M287_86nM244_334}
% \end{figure}

\item When we restrict the number of samples $\geq$ 25, the SVM shows a good recognition accuracy, 95.58\% over 12 classes. Again the accuracy of SVM also goes up compared to the SVM with \# of samples $\geq 12$. It implies that the more samples, the classifiers' performance improves.

\item SVM can cover 54 motion classes when  \# of samples $\geq 12$ with an accuracy 84.86\%. So, we use this trained SVM during {\em Adavu} recognition.

% It is yet to be tested that how those classifiers perform with more number of motion classes.
\end{itemize}

\section{{\em Adavu} Recognition} \label{sec:EditDistance}
To recognize an {\em Adavu}, we need to recognize the KPs and motions involved. The recognized sequence (i.e., KPs and motions) is matched with the dictionary of the {\em Adavu}s using Edit Distance algorithm. This algorithm computes the best match and recognizes the {\em Adavu}. Figure~\ref{fig:AdavuRecogworkflow} shows the workflow. In this, we use the trained models to recognise the KP (Using CNN) and motion (using SVM) as discussed in Section~\ref{sec:KPRecog} and \ref{sec:MotionRecog}.

First, we need to create a dictionary that keeps the ground truth of the {\em Adavu}~\cite{AdavuFile}. This ground truth is nothing but the labeling of an {\em Adavu} and the occurrence of KPs and motion IDs involved in that. This dictionary is used in Edit Distance algorithm to predict the best match.

\subsection{Result Discussion} \label{sec:ResultEditDistance}
This section discusses the prediction accuracy using Edit Distance algorithm. The inputs to this algorithm are an unknown sequence, and a dictionary denotes the {\em Adavu sequences} labeled with the {\em Adavu} name. The algorithm computes the distance score of the given unknown sequence with each of the {\em Adavu} sequences in the dictionary. The least score is treated as the best match. If it gives the same distance score for multiple {\em Adavu} sequences in the dictionary of an unknown sequence, then the first matched sequence's label name in the dictionary is given to the unknown one. We try two approaches for the {\em Adavu } recognition.

   {\bf Approach-1:} In this, only KP sequences are considered to recognise the {\em Adavu} as in ~\cite{sharma2013recognising} ~\cite{mallick2022posture} ~\cite{himadri2021} ~\cite{kale2015bharatna}.
   
    {\bf Approach-2:} For a given {\em Adavu}, the motions that do not participate in the recognition are predicted correctly in the given sequence of an {\em Adavu}. It is because, out of 334 motions, we consider only 54 during the recognition due to the lack of data. With this approach, we can include all the {\em Adavu}s during the {\em Adavu} recognition. So, a total 1645 number of sequences need to be predicted in this approach (Refer Table~\ref{tab:NumberofSequnces}).

In {\bf Approach-1}, 1275 sequences are predicted correctly out of 1645. That is, the recognition accuracy is 77.51\%. In this, the wrongly predicted 370 sequences belong to {\em Tatta\_2 to Tatta\_8} (For sequence count refer Table~\ref{tab:NumberofSequnces}). These are misclassified with {\em Tatta\_1} since the sequence of KPs in all these {\em Tatta Adavu}s are the same (Refer ~\cite{AdavuFile}). The KP sequence of all these {\em Tatta}: [P153 -- P153 -- P153 -- P153 -- .....-- P153]. As we can see, these {\em Adavu}s only comprises one KPs. Hence, the cross misclassifications occur within the {\em Tatta}. By considering only {\em Natta} and {\em Mettu Adavu}s as considered in ~\cite{himadri2021}, we achieve 100\% accuracy.

 {\bf Approach-2} attends an accuracy of 98.66\%. Here, out of 1645 {\em Adavu} sequences, 1623 predicted correctly. Here, 22 sequences of {\em Tatta-8 Adavu} are mispredicted as {\em Tatta-1} since These two {\em Adavu}s share common KPs and motions which follow the same order (Refer ~\cite{AdavuFile}). The sequence of {\em Tatta-1} and 8 is: [ M257 -- P153 -- M258 -- P153 -- M257 -- P153 -- M258 -- P153 -- M257 -- P153 -- M258 -- P153 -- M257 -- P153 -- M258 -- P153]. These two {\em Adavu}s are performed by two different {\em Sollukattu}s (Refer~\cite{mallick2017Data}). So they only can be distinguished by the audio beats, not by the video analysis.

 In both of these approaches, except {\em Tatta} the rest of the {\em Adavu}s' accuracy is 100\%. Though, in some cases, the KPs or motions may get mispredicted in an {\em Adavu}, the {\em Adavu} recognizer is still able to predict the correct {\em Adavu}. It is because, none of the motions or KPs are repeated in higher density (Refer ~\cite{AdavuFile}) in any of the {\em Adavu} sequences (except {\em Tatta}).

 Due to the non-overlapping sequences, the cross misclassifications are taken care by Edit Distance algorithm in both approaches. Now, let us take an example to understand this. The example is based on only KP sequences, and the same applies to KP and motions. Table~\ref{tab:Mettu_Kps_sequence} shows the sequence of KPs (ground truth sequences) involved in {\em Mettu Adavu}s. In these four {\em Mettu}s, all the KPs are unique, except P12 and P9. For {\em Mettu-1}, CNN predicts the KP sequence as \\
 {\small P2--\textcolor{red} {\underline{P1}}--P4--P1--P5--P6--P4--P1--P7--P8--P4--\textcolor{red}{\underline{P8}}--P5--P6--P4--\textcolor{red}{\underline{P3}}}\\
 Here, \textcolor{red}{\underline{red}} shows mispredicted KPs. When this predicted sequence is compared with the ground truth (Table~\ref{tab:Mettu_Kps_sequence}), Edit Distance algorithm yields distance values 3, 16, 16, and 16 with respect to {\em Mettu-1}, 2, 3, and 4. So, the predicted sequence matches with {\em Mettu-1}, even with three KP misclassifications. 
\begin{table}[!ht]
\centering
\scriptsize
\caption{Ground Truth KP Sequences in {\em Mettu}s \label{tab:Mettu_Kps_sequence}}
\small
\begin{tabular}{cl} \hline
{\bf \em Mettu} &\multicolumn{1}{c}{\bf KPs} \\ \hline
1 &P2--P3--P4--P1--P5--P6--P4--P1--P7--P8--P4--P1--P5--P6--P4--P1 \\ \hline
2 &P10--P11--P12--P9--P10--P11--P12--P9--P10--P11--P12--P9--P10--P11--P12--P9 \\ \hline
3 &P14--P15--P16--P13--P14--P15--P16--P13--P18--P19--P20--P17--P18--P19--P20--P17 \\ \hline

4 &P21--P22--P12--P9--P23--P24--P12--P9--P25--P26--P27--P28--P29--P30--P31--P32\\ \hline
\end{tabular}
\end{table} 
 
 Table~\ref{tab:ResultCompare} shows the result comparison of the contemporary approaches. As we can observe that the current approaches outperform the earlier ones, ~\cite{himadri2021}, ~\cite{mallick2022posture}, and ~\cite{sharma2013recognising}.
 \begin{table}[!ht]
 \scriptsize
     \centering
      \caption{Adavu Recognition Result Comparison}
     \begin{tabular}{c|p{1.2cm}|p{1.0cm}|c |c |c} \hline
          Approach & \multicolumn{2}{c|}{{\em Adavu} Attributes} &\# of & \# of  & Accuracy  \\  \cline{2-3}
          & KP & Motion & {\em Adavu}s & Sequences & (\%) \\ \hline
          SVM \& ED~\cite{himadri2021}  & \checkmark & -- &12 & 326 & 99.38 \\ \hline
          HMM  ~\cite{mallick2022posture} & \checkmark & -- & 8 & 56 & 94.64 \\ \hline
          HMM ~\cite{sharma2013recognising} & \checkmark & -- & 12 & -- & 80.55 \\ \hline
          
          CNN \& ED * & \checkmark& -- & 12 & 447 & 100 \\ 
          (Approach-1) & & & & & \\ \hline
          CNN, SVM \& ED* & \checkmark& \checkmark & 51 & 1645 & 98.66 \\ 
          (Approach-2) & & & & & \\ \hline
          \multicolumn{6}{c}{* mark denotes the proposed approaches}\\
          %\multicolumn{6}{c}{} \\
     \end{tabular}
     \begin{tabular}{lc|c|c}\\
     \multicolumn{4}{c}{\bf Execution time Comparison -- \cite{himadri2021} Vs Proposed Approaches} \\ \hline
         {Time}			& {SVM \& ED}\cite{himadri2021}	&{Approach-1} &{ Approach-2} \\ \hline
%Model Building Time ($CrT$)	&--- (s)	&--- (s)	&--- (s) \\ \hline
Prediction time ($P_rT$)	&21182 (s)	&278 (s)	 	&519 (s) \\ \hline
Average Prediction Time  ($AvgP_rT$)	&12.88 (s)	&0.169 (s)	&0.316 (s) \\ \hline
    \end{tabular}
     \label{tab:ResultCompare}
 \end{table}

\subsection{Performance Analysis Based on Time} 
We compute and compare the time complexity (Execution time) between the current approaches and earlier work ~\cite{himadri2021}. Table~\ref{tab:ResultCompare} shows the comparison. Here, the prediction time($P_rT$) implies the total time incurred to recognize all the {\em Adavu} sequences. Similarly, average prediction time ($AvgP_rT = P_rT/\gamma$) is the approximate prediction time per sequence. Here, The number of Sequences $\gamma=1645$, Refer Table~\ref{tab:NumberofSequnces}. The training time or model building time for SVM-KP, CNN-KP, and SVM-Motion recognition are 5121(s), 5472(s), and 1788(s), respectively. We carry the experiment on a system with Windows-10 OS, Intel Xeon Silver 4110 CPU, 32GB DDR4 RAM, and NVIDIA GeForce RTX 2080 Ti GPU with 11GB of GDDR6 RAM. In this system, Approach-1 is $\approx 76$ times faster than ~\cite{himadri2021}, whereas Approach-2, is $\approx 40$ times faster. While comparing the performance of the two proposed approaches, we find that Approach-1 is  $\approx 2$ times faster than Approach-2.

\section{Conclusion}\label{sec:conclusion}
The proposed approach recognizes the {\em Adavu} based on KP and motion rather than considering only KPs as in  ~\cite{himadri2021} ~\cite{mallick2022posture} ~\cite{sharma2013recognising}. This is the novelty of the work. Again, most of the earlier works use classical machine learning algorithms, whereas we use an effective deep learning technique, which gives a comparatively better result in less time. However, the robustness of the proposed CNN model is yet to be tested with the other data sets. The recognition accuracy (98.66\%) of the proposed one outperforms the previous approaches looking at the quantum of data. Though we try more sequences, we still achieve good accuracy. Again, the KP recognition uses CNN which is better than the SVM as reported in ~\cite{himadri2021} ~\cite{mallick2022posture} ~\cite{sharma2013recognising}. Moreover, the paper also explores and analyzes the performance of the proposed approaches based on prediction time. The same is also compared with the earlier similar work ~\cite{himadri2021} and found that the performance/efficiency of the proposed work is better than the earlier ones. For future work, we may try RNN for this sequence recognition since the occurrence of KPs and motions are the time series data, and RNN can handle this. In RNN, we can directly train the sequences (comprises of KPs and MHIs) and test the same for recognition.
%In this work, we create the trained models for KP and motion recognition. KP recognition uses CNN whereas for motion, we use SVM as classifier.  The paper also compare current approach with the contemporary results
\bibliographystyle{splncs04}
\bibliography{AdavuRecogRef}
\end{document}